\documentclass[letterpaper, 10 pt, conference]{ieeeconf}  
\IEEEoverridecommandlockouts                              
\usepackage{ps}
\setboolean{isTwoColumns}{false}

\preprinttrue
\publishedfalse

\glsdisablehyper 
\newcommand{\anAgent}{\ensuremath{i}}
\newcommand{\anotherAgent}{\ensuremath{j}}
\newcommand{\timeStep}{\ensuremath{t}}

\newcommand{\ofATimeStep}{\ensuremath{_{\timeStep}}}
\newcommand{\ofNextTimeStep}{\ensuremath{_{\timeStep+1}}}

\newcommand{\ofAnAgent}{\ensuremath{^{(\anAgent)}}}
\newcommand{\ofAnotherAgent}{\ensuremath{^{(\anotherAgent)}}}
\newcommand{\ofOtherAgents}{\ensuremath{^{(-\anAgent)}}}
\newcommand{\ofAgent}[1]{\ensuremath{^{(#1)}}}

\newglossaryentry{matrix:Adjacency}{
	name=\ensuremath{\bm{D}},
	description={Adjacency matrix},
	sort={D},
    type=symbol
}
\newcommand{\matAdjacency}{\gls{matrix:Adjacency}}
\newcommand{\matAdjacencyElement}[1]{\glslink{matrix:Adjacency}{\matAdjacency_{#1}}}

\newglossaryentry{set:realNumbers}{
	name=\ensuremath{\mathbb{R}},
	description={Set of real numbers},
	sort={real numbers},
    type=symbol
}
\newcommand{\setRealNumbers}{\gls{set:realNumbers}}

\newglossaryentry{set:naturalNumbers}{
	name=\ensuremath{\mathbb{N}},
	description={Set of natural numbers},
	sort={natural numbers},
    type=symbol
}
\newcommand{\setNaturalNumbers}{\gls{set:naturalNumbers}}

\newglossaryentry{set:systemStates}{
	name=\ensuremath{\mathcal{S}},
	description={Set of system states},
	sort={set of system states},
    type=symbol
}
\newcommand{\setSystemStates}{\gls{set:systemStates}}

\newglossaryentry{set:bigO}{
	name=\ensuremath{O},
	description={Big O},
	sort={O},
    type=symbol
}

\newglossaryentry{scalar:Weight}{
	name=\ensuremath{w},
	description={Weight},
	sort={weight},
    type=symbol
}

\newglossaryentry{scalar:NumberOfAgents}{
    name=\ensuremath{N},
    description={Number of agents},
    sort={Number of agents},
    type=symbol
}
\newcommand{\numAgents}{\gls{scalar:NumberOfAgents}}

\newglossaryentry{graph:path}{
    name=\ensuremath{\pi},
    description={Path},
    sort={Path},
    type=symbol
}
\newcommand{\graphPath}{\gls{graph:path}}

\newglossaryentry{scalar:NumberOfVerticesInPath}{
    name=\ensuremath{N_{\graphPath}},
    description={Number of vertices in path $\graphPath$, or length},
    sort={Number of vertices in path},
    type=symbol
}
\newcommand{\numVerticesPath}{\gls{scalar:NumberOfVerticesInPath}}

\newglossaryentry{trajectory:Reference}{
    name=\ensuremath{\bm{r}},
    description={Reference trajectory},
    sort={Reference Trajectory},
    type=symbol
}

\newglossaryentry{sym:horizonControl}{
	name=\ensuremath{N_u},
	description={Control horizon in model predictive control},
	sort={Nu},
    type=symbol
}

\newglossaryentry{sym:horizonPrediction}{
	name=\ensuremath{N_p},
	description={Prediction horizon in model predictive control},
	sort={Np},
    type=symbol
}

\newglossaryentry{sym:vehicleOrientation}{
	name=\ensuremath{\psi},
	description={Vehicle orientation},
	sort={psi},
    type=symbol
}

\newglossaryentry{sym:sysModelContinuous}{
    name=\ensuremath{f},
    description={Continuous-time system model},
    sort={f continuous-time},
    type=symbol
}
\newcommand{\sysModelContinuous}{\gls{sym:sysModelContinuous}}

\newglossaryentry{sym:sysModelDiscrete}{
    name=\ensuremath{f_{d}},
    description={Discrete-time system model},
    sort={f discrete-time},
    type=symbol
}

\newglossaryentry{sym:sysControlInputs}{
	name=\ensuremath{\bm{u}},
	description={System control inputs},
	sort=u,
    type=symbol
}
\NewDocumentCommand{\sysControlInputs}{ o }{\glslink{sym:sysControlInputs}{%
    \IfNoValueTF{#1}%
        {\ensuremath{\bm{u}}}%
        {\ensuremath{\bm{u}^{(#1)}}}%
}}

\newglossaryentry{sym:outputs}{
	name=\ensuremath{\bm{y}},
	description={System outputs},
	sort={y},
    type=symbol
}

\newglossaryentry{sym:sysSpeed}{
	name=\ensuremath{\mathrm{v}},
	description={Vehicle speed},
	sort={v},
    type=symbol
}
\newcommand{\sysSpeed}{\gls{sym:sysSpeed}}

\newglossaryentry{sym:inSpeed}{
	name=\ensuremath{u_{\sysSpeed}},
	description={Vehicle input speed},
	sort={uv},
    type=symbol
}

\newglossaryentry{sym:steering-angle}{
	name=\ensuremath{\delta},
	description={Vehicle steering angle},
	sort={delta},
    type=symbol
}

\newglossaryentry{sym:inSteering}{
	name=\ensuremath{u_{\delta}},
	description={Vehicle input steering angle},
	sort={ud},
    type=symbol
}

\newglossaryentry{sym:nColors}{
	name=\ensuremath{N_c},
	description={Number of colors},
	sort={Number of colors},
    type=symbol
}

\newglossaryentry{sym:nStates}{
	name=\ensuremath{n},
	description={Number of states of a dynamical system},
	sort={Number of states},
    type=symbol
}
\newcommand{\numStates}{\gls{sym:nStates}}

\newglossaryentry{sym:nInputs}{
    name=\ensuremath{m},
    description={Number of inputs of a dynamical system},
    sort={m number of inputs},
    type=symbol
}
\newcommand{\numInputs}{\gls{sym:nInputs}}

\newglossaryentry{sym:nLevels}{
	name=\ensuremath{N_{\text{CL}}},
	description={Number of computation levels},
	sort={Number of computation levels},
    type=symbol
}

\newglossaryentry{sym:nLevelsAllowed}{
	name=\ensuremath{N_{\text{CL},a}},
	description={Allowed number of computation levels},
	sort={Number of computation levels allowed},
    type=symbol
}

\newglossaryentry{sym:numGroups}{
	name=\ensuremath{N_{g}},
	description={Number of parallelly computing groups of agents},
	sort={Number of groups},
    type=symbol
}

\newglossaryentry{sym:fnPrio}{
    name=\ensuremath{p},
    description={Priority assignment function},
    sort={Priority assignment function},
    type=symbol
}

\newglossaryentry{sym:tSample}{
	name=\ensuremath{T_s},
	description={Sample Time},
	sort={T sample},
    type=symbol
}

\newglossaryentry{sym:tSolve}{
	name=\ensuremath{T_\text{sol.}},
	description={Computation time \tSolveB{\anAgent} that agent $\anAgent$ needs to solve its \ac{ocp}},
	sort={T solve},
    type=symbol
}

\newcommand{\tSolveB}[1]{\glslink{sym:tSolve}{\ensuremath{\ensuremath{T_\text{sol.}}^{(#1)}}}}

\newglossaryentry{sym:tSolveUpper}{
	name=\ensuremath{T_\text{sol.,max}},
	description={Upper computation time $T_\text{sol.,max}\ofAgent{\anAgent}$ that agent $\anAgent$ needs to solve it \ac{ocp}},
	sort={T solve upper},
    type=symbol
}

\newglossaryentry{sym:vertices}{
	name=\ensuremath{\mathcal{V}},
	description={Set of vertices},
	sort={Vertices},
    type=symbol
}
\newcommand{\setVertices}{\gls{sym:vertices}}
\newcommand{\setAgents}{\setVertices}

\newcommand{\helpSetPredecessors}[1]{\ensuremath{\setVertices^{(#1\leftarrow)}}}
\newglossaryentry{sym:predecessors}{
	name=\ensuremath{\helpSetPredecessors{i}},
	description={Set of predecessors of vertex $i$},
	sort={Vertices 1},
    type=symbol
}
\newcommand{\setPredecessors}[1]{\glslink{sym:predecessors}{\ensuremath{\helpSetPredecessors{#1}}}}

\newcommand{\helpSetPredecessorsPar}[1]{\ensuremath{\setVertices^{(#1\leftarrow)}_{\text{par.}}}}
\newglossaryentry{sym:predecessorsPar}{
	name=\ensuremath{\helpSetPredecessorsPar{i}},
	description={Set of predecessors of vertex $i$ that have parallel couplings with it},
	sort={Vertices 2},
    type=symbol
}

\newcommand{\helpSetPredecessorsSeq}[1]{\ensuremath{\setVertices^{(#1\leftarrow)}_{\text{seq.}}}}
\newglossaryentry{sym:predecessorsSeq}{
	name=\ensuremath{\helpSetPredecessorsSeq{i}},
	description={Set of predecessors of vertex $i$ that have sequential couplings with it},
	sort={Vertices 3},
    type=symbol
}

\newcommand{\helpSetSuccessors}[1]{\ensuremath{\setVertices^{(#1\rightarrow)}}}
\newglossaryentry{sym:successors}{
	name=\ensuremath{\helpSetSuccessors{i}},
	description={Set of successors of vertex $i$},
	sort={Vertices 4},
    type=symbol
}
\newcommand{\setSuccessors}[1]{\glslink{sym:successors}{\ensuremath{\helpSetSuccessors{#1}}}}

\newglossaryentry{sym:neighbors}{
	name=\ensuremath{\setVertices^{(i)}},
	description={Set of neighbors of vertex $i$},
	sort={Vertices 0},
    type=symbol
}
\newcommand{\setNeighbors}[1]{\glslink{sym:neighbors}{\ensuremath{\setVertices^{(#1)}}}}

\newglossaryentry{sym:degree}{
	name=\ensuremath{d^{(i)}},
	description={Degree of vertex $i$. Sum of in-degree and out-degree},
	sort=degree,
    type=symbol
}
\newcommand{\vertexDegree}[1]{\glslink{sym:degree}{\ensuremath{d^{(#1)}}}}

\newcommand{\helpVertexInDegree}[1]{\ensuremath{d^{(#1\leftarrow)}}}
\newglossaryentry{sym:inDegree}{
    name=\helpVertexInDegree{i},
    description={In-degree of vertex $i$},
    sort={degree in},
    type=symbol
}
\newcommand{\vertexInDegree}[1]{\glslink{sym:inDegree}{\helpVertexInDegree{#1}}}

\newcommand{\helpVertexOutDegree}[1]{\ensuremath{d^{(#1\rightarrow)}}}
\newglossaryentry{sym:outDegree}{
    name=\helpVertexOutDegree{i},
    description={Out-degree of vertex $i$},
    sort={degree out},
    type=symbol,
}
\newcommand{\vertexOutDegree}[1]{\glslink{sym:outDegree}{\helpVertexOutDegree{#1}}}

\newglossaryentry{sym:matLevels}{
	name=\ensuremath{\bm{L}},
	description={Matrix of computation levels},
	sort=L,
    type=symbol
}

\newglossaryentry{sym:tComp}{
	name=\ensuremath{T},
	description={Computation time},
	sort={T},
    type=symbol
}

\newglossaryentry{sym:tCompNcs}{
	name=\ensuremath{T_{\text{NCS}}},
	description={Computation time of \iac{ncs}},
	sort={T NCS},
    type=symbol
}
\newcommand{\tCompNcs}{\gls{sym:tCompNcs}}

\newglossaryentry{graph:Undirected}{
	name=\ensuremath{\mathcal{G}},
	description={Undirected Graph},
	sort={graph1},
    type=symbol
}
\newcommand{\graphUndirected}{\gls{graph:Undirected}}

\newglossaryentry{graph:Directed}{
    name=\ensuremath{\vec{\gls*{graph:Undirected}}},
	description={Directed Graph},
	sort={graph2},
    type=symbol
}
\newcommand{\graphDirected}{\gls{graph:Directed}}

\newglossaryentry{mat:edgeUtilities}{
    name=\ensuremath{M_\text{u}},
	description={Edge utility matrix},
	sort={matrix edge utilities},
    type=symbol
}

\newglossaryentry{sym:setColors}{
    name=\ensuremath{\mathcal{C}},
    description={Set of colors},
    sort=Colors,
    type=symbol
}

\newglossaryentry{sym:varControlInvariantSet}{
	name=\ensuremath{\mathcal{C}_{\text{inv}}},
	description={Control invariant set},
	sort={Control invariant set},
    type=symbol
}

\newglossaryentry{set:Weights}{
	name=\ensuremath{\mathcal{W}},
	description={Set of weights in a weighted graph},
    sort={Weights},
    type=symbol
}

\newglossaryentry{set:Edges}{
	name=\ensuremath{\mathcal{E}},
	description={Set of edges; used to indicate that only undirected edges exist},
    sort={Edges},
    type=symbol
}
\newcommand{\setEdges}{\gls{set:Edges}}

\newglossaryentry{sym:setEdgesDirected}{
	name=\ensuremath{\vec{\gls*{set:Edges}}},
	description={Set of directed edges},
	sort={Edges directed},
    type=symbol
}

\newglossaryentry{sym:varEdge}{
	name=\ensuremath{(i \rightarrow j)},
	description={Directed edge from vertex $i$ to vertex $j$},
	sort={edge},
    type=symbol
}
\newcommand{\edgeDirected}[2]{\glslink{sym:varEdge}{\ensuremath{(#1 \rightarrow #2)}}}

\newglossaryentry{sym:fnReorder}{
	name=\ensuremath{f_r},
	description={Reordering function for graph color values},
	sort=fr,
    type=symbol
}

\newglossaryentry{sym:fcnObjective}{
    name=\ensuremath{J},
    description={Objective function of an optimization problem},
    sort=J,
    type=symbol
}
\NewDocumentCommand{\fcnObjective}{ o }{\glslink{sym:fcnObjective}{%
    \IfNoValueTF{#1}%
        {\ensuremath{J}}%
        {\ensuremath{J^{(#1)}}}%
}}

\newglossaryentry{sym:fcnObjectiveState}{
    name=\ensuremath{\ell_{x}},
    description={Reference tracking objective function},
    sort={lx Reference tracking objective function},
    type=symbol
}
\NewDocumentCommand{\fcnObjectiveState}{ o }{\glslink{sym:fcnObjectiveState}{%
    \IfNoValueTF{#1}%
        {\ensuremath{\ell_{x}}}%
        {\ensuremath{\ell_{x}^{(#1)}}}%
}}

\newglossaryentry{sym:fcnObjectiveStateTerminal}{
    name=\ensuremath{\ell_{f}},
    description={Reference tracking objective terminal function},
    sort={lf Reference tracking objective terminal function},
    type=symbol
}

\newglossaryentry{sym:fcnObjectiveInput}{
    name=\ensuremath{\ell_{u}},
    description={Input change objective function},
    sort={lu Input change objective function},
    type=symbol
}
\NewDocumentCommand{\fcnObjectiveInput}{ o }{\glslink{sym:fcnObjectiveInput}{%
    \IfNoValueTF{#1}%
        {\ensuremath{\ell_{u}}}%
        {\ensuremath{\ell_{u}^{(#1)}}}%
}}

\newglossaryentry{sym:fcnObjectiveCoupling}{
    name=\ensuremath{\ell_\text{c}},
    description={Coupling objective function},
    sort={lc Coupling objective function},
    type=symbol
}
\NewDocumentCommand{\fcnObjectiveCoupling}{ oo }{\glslink{sym:fcnObjectiveCoupling}{%
    \IfNoValueTF{#1}%
        {\ensuremath{\ell_\text{c}}}%
        {\ensuremath{\ell_\text{c}^{(#1,#2)}}}%
}}

\newglossaryentry{sym:fcnConstraintCoupling}{
    name=\ensuremath{c_\text{c}},
    description={Coupling constraint function},
    sort={cc Coupling constraint function},
    type=symbol
}
\NewDocumentCommand{\fcnConstraintCoupling}{ oo }{\glslink{sym:fcnConstraintCoupling}{%
    \IfNoValueTF{#1}%
        {\ensuremath{c_\text{c}}}%
        {\ensuremath{c_\text{c}^{(#1,#2)}}}%
}}

\newglossaryentry{sym:prediction}{
	name=\ensuremath{\tilde{\bm{x}}^{(j \leftarrow i)}_{\cdot \vert k}},
	description={Prediction in agent $i$ for agent $j$ at time $k$},
	sort=x,
    type=symbol,
}
\newcommand{\agentPrediction}{\glslink{sym:prediction}{\ensuremath{ \tilde{\bm{x}} }}}

\newcommand{\agentPredictionForAgentAFromAgentBAtTimeC}[3]{\glslink{sym:prediction}{\ensuremath{ \agentPrediction^{(#1 \leftarrow #2)}_{\cdot \vert #3} }}}

\newglossaryentry{sym:state}{
	name=\ensuremath{\bm{x}},
	description={System state},
	sort=x,
    type=symbol
}
\newcommand{\sysState}{\gls{sym:state}}
\newglossaryentry{sym:ref}{
	name=\ensuremath{\bm{r}},
	description={System reference},
	sort=r,
    type=symbol
}

\newglossaryentry{sym:stateAgent}{
	name=\ensuremath{\sysState^{(i)}_{(k)}},
	description={System state of agent $i$ at time $k$},
	sort=x,
    type=symbol,
}

\newglossaryentry{sym:setReachable}{
	name=\ensuremath{\mathcal{R}^{(i)}},
	description={reachable set of agent $i$},
	sort={Reachable set},
    type=symbol
}
\newcommand{\setReachable}{\glslink{sym:setReachable}{\ensuremath{\mathcal{R}}}}

\newglossaryentry{set:occupiedArea}{
	name=\ensuremath{\mathcal{O}^{(i)}},
	description={Set of the occupied area of the predicted trajectory of agent $\anAgent$},
	sort={occupied area},
    type=symbol
}

\newglossaryentry{set:feasibleStates}{
	name=\ensuremath{\mathcal{X}},
	description={set of feasible states},
	sort={x},
    type=symbol
}
\newcommand{\setFeasibleStates}{\gls{set:feasibleStates}}

\newglossaryentry{set:feasibleInputs}{
	name=\ensuremath{\mathcal{U}},
	description={set of feasible inputs},
	sort={u},
    type=symbol
}
\newcommand{\setFeasibleInputs}{\gls{set:feasibleInputs}}

\newglossaryentry{sym:numStatesConfSpace}{
    name=\ensuremath{n_p},
    description={Number of states that are in the conflictual space},
    sort={n number of states that are in the conflictual space},
    type=symbol
}

\newglossaryentry{sym:fcnProj}{
    name=\text{proj},
    description={A function that projects a reachable set of system states in the conflictual space},
    sort={Project function},
    type=symbol
}

\newglossaryentry{rl:setOfAgents}{
    name=\ensuremath{\mathcal{N}},
    description={A set of agents},
    sort={Set of agents},
    type=symbol
}
\newcommand{\setOfAgents}{\gls{rl:setOfAgents}}

\newglossaryentry{rl:actionSpace}{
    name=\ensuremath{\mathcal{A}},
    description={Action space},
    sort={Action space},
    type=symbol
}
\newcommand{\actionSpace}{\gls{rl:actionSpace}}

\newglossaryentry{rl:stateSpace}{
    name=\ensuremath{\mathcal{S}},
    description={State space},
    sort={State space},
    type=symbol
}
\newcommand{\stateSpace}{\gls{rl:stateSpace}}

\newglossaryentry{rl:observationSpace}{
    name=\ensuremath{\mathcal{O}},
    description={Observation space},
    sort={Observation space},
    type=symbol
}
\newcommand{\observationSpace}{\gls{rl:observationSpace}}

\newglossaryentry{rl:policySpace}{
    name=\ensuremath{\Omega},
    description={Policy space},
    sort={Policy space},
    type=symbol
}
\newcommand{\policySpace}{\gls{rl:policySpace}}

\newglossaryentry{rl:observationFcn}{
    name=\ensuremath{\Omega},
    description={Observation function},
    sort={Observation function},
    type=symbol
}

\newglossaryentry{rl:transitionProbFcn}{
    name=\ensuremath{\mathcal{P}},
    description={Transition probability function},
    sort={Transition probability function},
    type=symbol
}
\newcommand{\transitionProbFcn}{\gls{rl:transitionProbFcn}}

\newglossaryentry{rl:rewardFcn}{
    name=\ensuremath{R},
    description={Reward function},
    sort={Reward function},
    type=symbol
}
\newcommand{\rewardFcn}{\gls{rl:rewardFcn}}

\newglossaryentry{rl:discountFactor}{
    name=\ensuremath{\gamma},
    description={Discount factor},
    sort={Discount factor},
    type=symbol
}
\newcommand{\discountFactor}{\gls{rl:discountFactor}}

\newglossaryentry{rl:state}{
    name=\ensuremath{s},
    description={State},
    sort={State},
    type=symbol
}
\newcommand{\state}{\gls{rl:state}}

\newglossaryentry{rl:nextState}{
    name=\ensuremath{s'},
    description={Next state},
    sort={Next state},
    type=symbol
}

\newglossaryentry{rl:action}{
    name=\ensuremath{a},
    description={Action},
    sort={Action},
    type=symbol
}
\newcommand{\action}{\gls{rl:action}}

\newglossaryentry{rl:jointActions}{
    name=\ensuremath{a},
    description={Joint actions},
    sort={Joint actions},
    type=symbol
}

\newglossaryentry{rl:observation}{
    name=\ensuremath{o},
    description={Observation},
    sort={Observation},
    type=symbol
}
\newcommand{\observation}{\gls{rl:observation}}

\newglossaryentry{rl:jointObservations}{
    name=\ensuremath{\bm{o}},
    description={Joint observations},
    sort={Joint observations},
    type=symbol
}

\newglossaryentry{rl:reward}{
    name=\ensuremath{r},
    description={Reward},
    sort={Reward},
    type=symbol
}

\newglossaryentry{rl:jointRewards}{
    name=\ensuremath{\bm{r}},
    description={Joint rewards},
    sort={Joint rewards},
    type=symbol
}

\newglossaryentry{rl:valueFunction}{
    name=\ensuremath{V},
    description={A function that evaluates how good a policy is},
    sort={Value function},
    type=symbol
}
\newcommand{\valueFunction}{\gls{rl:valueFunction}}

\newglossaryentry{rl:policy}{
    name=\ensuremath{\pi},
    description={Policy},
    sort={Policy},
    type=symbol
}
\newcommand{\policy}{\glslink{rl:policy}{\ensuremath{\pi}}}

\newglossaryentry{rl:policyOptimal}{
    name=\ensuremath{\pi^{*}{}},
    description={Optimal policy},
    sort={Optimal policy},
    type=symbol
}
\newcommand{\policyOptimal}{\gls{rl:policyOptimal}}

\newcommand{\baseSubscript}{\textbf{D}} 
\newcommand{\priSubscript}{\textbf{P}} 

\newglossaryentry{rl:setOfPolicies}{
    name=\ensuremath{\Pi},
    description={Set of policies},
    sort={Set of policies},
    type=symbol
}
\newcommand{\setOfPolicies}{\gls{rl:setOfPolicies}}

\newglossaryentry{rl:setOfPoliciesBase}{
    name=\ensuremath{\setOfPolicies_{\baseSubscript}},
    description={Set of policies of the decision-making policy},
    sort={Set of policies of the decision-making policy},
    type=symbol
}

\newglossaryentry{rl:setOfPoliciesPri}{
    name=\ensuremath{\setOfPolicies_{\priSubscript}},
    description={Set of policies of the priority-assignment policy},
    sort={Set of policies of the priority-assignment policy},
    type=symbol
}

\newglossaryentry{rl:setOfActions}{
    name=\ensuremath{A},
    description={Set of actions},
    sort={Set of actions},
    type=symbol
}
\newcommand{\setOfActions}{\gls{rl:setOfActions}}

\newglossaryentry{rl:setOfActionsBase}{
    name=\ensuremath{\setOfActions_{\baseSubscript}},
    description={Set of actions of the decision-making policy},
    sort={Set of actions of the decision-making policy},
    type=symbol
}

\newglossaryentry{rl:setOfActionsPri}{
    name=\ensuremath{\setOfActions_{\priSubscript}},
    description={Set of actions of the priority-assignment policy},
    sort={Set of actions of the priority-assignment policy},
    type=symbol
}

\newglossaryentry{rl:setOfObservations}{
    name=\ensuremath{O},
    description={Set of observations},
    sort={Set of observations},
    type=symbol
}
\newcommand{\setOfObservations}{\gls{rl:setOfObservations}}

\newglossaryentry{rl:setOfObservationsBase}{
    name=\ensuremath{\setOfObservations_{\baseSubscript}},
    description={Set of observations of the decision-making policy},
    sort={Set of observations of the decision-making policy},
    type=symbol
}

\newglossaryentry{rl:setOfObservationsPri}{
    name=\ensuremath{\setOfObservations_{\priSubscript}},
    description={Set of observations of the priority-assignment policy},
    sort={Set of observations of the priority-assignment policy},
    type=symbol
}

\newglossaryentry{sym:distance}{
    name=\ensuremath{d},
    description={Distance},
    sort={Distance},
    type=symbol
}

\newglossaryentry{sym:numOfPointsRef}{
    name=\ensuremath{n_\text{p,RP}},
    description={Number of points on the reference path},
    sort={Number of points on the reference path},
    type=symbol
}

\newglossaryentry{sym:numOfObservedSurroundingAgents}{
    name=\ensuremath{n_\text{sur.}},
    description={Number of observed surrounding agents},
    sort={Number of observed surrounding agents},
    type=symbol
}

\newglossaryentry{sym:numOfNotMaskedSurroundingAgents}{
    name=\ensuremath{n_\text{sur.,NM}},
    description={Number of not masked surrounding agents},
    sort={Number of not masked agents},
    type=symbol
}

\newglossaryentry{sym:numOfMaskedSurroundingAgents}{
    name=\ensuremath{n_\text{sur.,M}},
    description={Number of masked surrounding agents},
    sort={Number of masked agents},
    type=symbol
}

\newglossaryentry{rl:numOfSamples}{
    name=\ensuremath{n_\text{samples}},
    description={Number of training samples},
    sort={Number of training samples},
    type=symbol
}

\newglossaryentry{rl:setOfModels}{
    name=\ensuremath{\mathcal{M}},
    description={A set of models},
    sort={A set of models},
    type=symbol
}

\newglossaryentry{rl:maxModelIndex}{
    name=\ensuremath{5},
    description={Maximum Model Index},
    sort={Maximum Model Index},
    type=symbol
}

\newglossaryentry{rl:numberOfModels}{
    name=\text{six},
    description={Number of models},
    sort={Number of models},
    type=symbol
}

\newglossaryentry{rl:model}{
    name=\ensuremath{M},
    description={RL Model},
    sort={RL Model},
    type=symbol
}
\newcommand{\model}{\gls{rl:model}}

\newglossaryentry{rl:compositeScore}{
    name=\ensuremath{CS},
    description={Composite Score},
    sort={Composite Score},
    type=symbol
}

\newglossaryentry{rl:collisionRate}{
    name=\ensuremath{CR},
    description={Collision Rate},
    sort={Collision Rate},
    type=symbol
}

\newglossaryentry{rl:collisionRateTotal}{
    name=\ensuremath{CR_{\text{total}}},
    description={Total Collision Rate},
    sort={Total Collision Rate},
    type=symbol
}

\newglossaryentry{rl:collisionRateAA}{
    name=\ensuremath{CR_{\text{A-A}}},
    description={Agent-Agent Collision Rate},
    sort={Agent-Agent Collision Rate},
    type=symbol
}

\newglossaryentry{rl:collisionRateAL}{
    name=\ensuremath{CR_{\text{A-L}}},
    description={Agent-Lanelet Collision Rate},
    sort={Agent-Lanelet Collision Rate},
    type=symbol
}

\newglossaryentry{rl:safetyRate}{
    name=\ensuremath{SR},
    description={Safety Rate},
    sort={Safety Rate},
    type=symbol
}

\newglossaryentry{rl:safetyRateTotal}{
    name=\ensuremath{SR_{\text{total}}},
    description={Total Safety Rate},
    sort={Total Safety Rate},
    type=symbol
}

\newglossaryentry{rl:safetyRateAA}{
    name=\ensuremath{SR_{\text{A-A}}},
    description={Agent-Agent Safety Rate},
    sort={Agent-Agent Safety Rate},
    type=symbol
}

\newglossaryentry{rl:safetyRateAL}{
    name=\ensuremath{SR_{\text{A-L}}},
    description={Agent-Lanelet Safety Rate},
    sort={Agent-Lanelet Safety Rate},
    type=symbol
}

\newglossaryentry{rl:centerLineDeviation}{
    name=\ensuremath{CD},
    description={Center Line Deviation},
    sort={Score: Center Line Deviation},
    type=symbol
}

\newglossaryentry{rl:laneAdherence}{
    name=\ensuremath{LA},
    description={Lane Adherence},
    sort={Score: Lane Adherence},
    type=symbol
}

\newglossaryentry{rl:averageSpeed}{
    name=\ensuremath{AS},
    description={Average Speed},
    sort={Average Speed},
    type=symbol
}

\newglossaryentry{rl:baseProblem}{
    name=\ensuremath{\mathcal{G}_{\baseSubscript}},
    description={Base problem},
    sort={Base problem},
    type=symbol
}

\newglossaryentry{rl:priProblem}{
    name=\ensuremath{\mathcal{G}_{\priSubscript}},
    description={Priority assignment problem},
    sort={Priority assignment problem},
    type=symbol
}

\newglossaryentry{rl:priRank}{
    name=\ensuremath{\mathcal{R}_{\priSubscript}},
    description={Priority rank},
    sort={Priority rank},
    type=symbol
}
\newcommand{\priRank}{\glslink{rl:priRank}{\ensuremath{\mathcal{R}_{\priSubscript}}}}
\newcommand{\priRankB}[1]{\glslink{rl:priRank}{\ensuremath{\mathcal{R}_{\priSubscript,#1}}}}

\newglossaryentry{rl:higherPriorities}{
    name=\ensuremath{\leftarrow},
    description={Higher priorities},
    sort={Higher priorities},
    type=symbol
}
\newcommand{\higherPriorities}{\gls{rl:higherPriorities}}

\newcommand{\setOfHigherPriorityAgents}[1]{\ensuremath{\setOfAgents^{(#1 \higherPriorities)}}}

\newcommand{\setOfObservableHigherPriorityAgents}[1]{\ensuremath{\setOfAgents^{(#1 \higherPriorities)}_{\text{obs.}}}}

\DeclareMathOperator*{\argsort}{arg\,sort}
\DeclareMathOperator*{\where}{where}

\newcommand{\frameworkName}{\text{XP-MARL}\xspace}
\DeclareAcronym{ap}{
    short = AP,
    long  = Action Propagation,
}
\DeclareAcronym{cav}{
    short = CAV,
    long  = Connected and Automated Vehicle,
}
\DeclareAcronym{cg}{
    short = CG,
    long = Center of Gravity,
    short-plural = s,
    long-plural-form = Centers of Gravity,
}
\DeclareAcronym{cnn}{
    short = CNN,
    long  = Convolutional Neural Network
}
\DeclareAcronym{cpm}{
    short = CPM,
    long  = Cyber-Physical Mobility
}
\DeclareAcronym{cpmlab}{
    short = CPM Lab,
    long  = Cyber-Physical Mobility Lab
}
\DeclareAcronym{dmpc}{
    short = DMPC,
    long  = distributed model predictive control
}
\DeclareAcronym{dql}{
    short = DQL,
    long  = Deep Q-Learning
}

\DeclareAcronym{il}{
    short = IL,
    long  = Imitation Learning,
    short-indefinite = an,
}
\DeclareAcronym{mappo}{
    short = MAPPO,
    long  = Multi-Agent \ac{ppo},
    short-indefinite = an,
}
\DeclareAcronym{maddpg}{
    short = MADDPG,
    long  = Multi-Agent Deep Deterministic Policy Gradient,
    short-indefinite = an,
}
\DeclareAcronym{mas}{
    short = MAS,
    long  = Multi-Agent System,
    short-indefinite = an,
}
\DeclareAcronym{mdp}{
    short = MDP,
    long  = Markov decision process,
    short-indefinite = an,
}
\DeclareAcronym{mg}{
    short = MG,
    long  = Markov Game,
    short-indefinite = an,
}
\DeclareAcronym{ml}{
    short = ML,
    long  = Machine Learning,
    short-indefinite = an,
}
\DeclareAcronym{mpc}{
    short = MPC,
    long  = model predictive control,
    short-indefinite = an,
}
\DeclareAcronym{marl}{
    short = MARL,
    long  = Multi-Agent Reinforcement Learning,
    short-indefinite = an,
}
\DeclareAcronym{ocp}{
    short = OCP,
    long  = optimal control problem,
    short-indefinite = an,
    long-indefinite = an,
}
\DeclareAcronym{per}{
    short = PER,
    long  = Prioritized Experience Replay
}
\DeclareAcronym{pomdp}{
    short = POMDP,
    long  = Partially Observable \ac{mdp}
}
\DeclareAcronym{pomg}{
    short = POMG,
    long  = Partially Observable \ac{mg}
}
\DeclareAcronym{ppo}{
    short = PPO,
    long  = Proximal Policy Optimization
}
\DeclareAcronym{rhc}{
    short = RHC,
    long  = receding horizon control,
    short-indefinite = an,
}
\DeclareAcronym{rl}{
    short = RL,
    long  = Reinforcement Learning,
    short-indefinite = a,
}
\DeclareAcronym{som}{
    short = SOM,
    long  = Self Other-Modeling,
    short-indefinite = a,
}
\DeclareAcronym{zsg}{
    short = ZSG,
    long  = Zero-Shot Generalization,
}
\newglossaryentry{def:agent}{
	name=agent,
	description={An agent is a system which is composed of at least one of the three elements: sensors, actuators, and a dynamic behavior.%
    },
}

\newglossaryentry{def:agentActive}{
	name=active agent,
	description={Active \glspl{def:agent} are \glspl{def:agent} which are connected using a communication
    network over which they can exchange data. The exchanged data is
    used by the \glspl{def:agent}' controllers to find appropriate inputs to achieve their
    goals while interacting with other \glspl{def:agent}.
    Additionally, active \glspl{def:agent} consider \glspl{def:agentPassive}},
    parent=def:agent,
}

\newglossaryentry{def:agentPassive}{
	name=passive agent,
	description={Passive \glspl{def:agent} are \glspl{def:agent} without networked control. However, they can communicate their data like current and future states to \glspl{def:agentActive}, or they can be detected by \glspl{def:agentActive}' sensors.%
    },
    parent=def:agent,
}

\newglossaryentry{def:distrutedSolutionQuality}{
	name=distributed solution quality,
	description={%
        The quality $q\in [0,1]$ of the solution in \ac{dmpc} is the networked objective function value ${\fcnObjective}_{c}$ for the solution of the corresponding \ac{cmpc} formulation divided by the objective function value ${\fcnObjective_d}$ for the solution of the \ac{dmpc} formulation
        \begin{equation}
            q = \frac{\fcnObjective_c}{\fcnObjective_d}.
        \end{equation}
    },
}

\newglossaryentry{def:mas}{
	name=multi-agent system,
	description={A system consisting of multiple \glspl{def:agent}.%
    },
}

\newglossaryentry{def:ncs}{
	name=networked control system,
	description={A system consisting of multiple \glspl{def:agentActive}.%
    },
}

\newglossaryentry{def:prediction}{
	name=prediction,
	description={
        A prediction $\agentPrediction^{\anAgent}_{\cdot\vert \timestep}$ of \gls{def:agent} $\anAgent$ is its predicted state trajectory as obtained from the solution of its \ac{ocp} at time $\timestep$.
        A prediction $\agentPredictionForAgentAFromAgentBAtTimeC{\anAgent}{\anotherAgent}{\timestep}$ of \gls{def:agent} $\anAgent$ for \gls{def:agent} $\anotherAgent$ is agent $\anotherAgent$'s state trajectory as viewed from agent $\anAgent$ at time $\timestep$. It is obtained by communication or by predicting \gls{def:agent} $\anotherAgent$'s state trajectory with its model using the solution to its \ac{ocp}.%
    },
}

\newglossaryentry{def:consistency}{
	name=prediction consistency,
	description={%
        \Iac{ncs} is prediction consistent at time step $\timestep$ if the \gls{def:prediction} \agentPredictionForAgentAFromAgentBAtTimeC{\anotherAgent}{\anAgent}{\timestep} of every agent $\anAgent\in\setAgents$ for each of its neighbors $\anotherAgent \in \setNeighbors{\anAgent}$ equals the actual \gls{def:prediction} $\agentPrediction^{(\anotherAgent)}_{\cdot\vert \timestep}$ of its neighbors, i.e.,
        \begin{equation}
            \agentPredictionForAgentAFromAgentBAtTimeC{\anotherAgent}{\anAgent}{\timestep}=\agentPrediction^{(\anotherAgent)}_{\cdot\vert \timestep}, \quad \forall \anAgent \in \setAgents, \forall \anotherAgent \in \setNeighbors{\anAgent}.
        \end{equation}%
    }
}

\newglossaryentry{def:ncsFeasible}{
	name=NCS-feasible,
	description={
        The solutions $\sysControlInputs_{\cdot \vert \timestep}\ofAgent{\anAgent}$ of all agents $i\in\setAgents$ are \acs*{ncs}-feasible if the stacked solution vector $\bm{U}_{\cdot \vert \timestep} = \left( \sysControlInputs_{\cdot \vert \timestep}\ofAgent{1}, \ldots, \sysControlInputs_{\cdot \vert \timestep}\ofAgent{\numAgents} \right)\transposed$ satisfies all constraints of the corresponding central \acf*{ocp} considering all agents.%
    },
}

\newglossaryentry{def:feasibleAgent}{
	name=agent-feasible,
	description={%
        A solution is agent-feasible if it satisfies the constraints of to the corresponding agent's \ac{ocp}.%
    },
}

\newglossaryentry{def:networkedObjectiveFunction}{
	name=networked objective function,
	description={%
        The objective function value ${\fcnObjective}$ in \iac{ncs} formulation is the sum of all agent objective functions \fcnObjective[i]
        \begin{equation}
            \fcnObjective = \sum_{i}^{i\in\setAgents} \fcnObjective[i].
        \end{equation}
    },
}

\newglossaryentry{def:optimalPriorityAssignment}{
	name=optimal priority assignment,
	description={%
        The optimal priority assignment results in a feasible solution for every agent with the lowest networked objective function value.%
    },
}

\newglossaryentry{def:graph}{
	name=graph,
	description={%
        A directed graph $\graphDirected = \left(\setVertices,\setEdges\right)$ is a pair of two sets,
        the set of vertices $\setVertices=\set{1,\dots,\numAgents}$
        and the set of directed edges $\setEdges \subseteq \setVertices \times \setVertices$.
        The edge from $i$ to $j$ is denoted by $\edgeDirected{i}{j}$.
        An undirected graph $\graphUndirected = \left(\setVertices,\setEdges\right)$ is a special form of a directed graph in which every edge is directed both ways, i.e., $\edgeDirected{i}{j} \in \setEdges \iff \edgeDirected{j}{i} \in \setEdges$.
    },
}

\newglossaryentry{def:path}{
	name=path,
	description={%
        A path of a graph $\graphDirected$ is a subgraph $\graphDirected_{\graphPath} = \left(\setVertices_{\graphPath},\setEdges_{\graphPath}\right)\subseteq\graphDirected$ with distinct vertices $\setVertices_{\graphPath}=\{i_{1},i_{2},i_{3},\ldots,i_{\numVerticesPath-1},i_{\numVerticesPath}\}$ and distinct edges $\setEdges_{\graphPath}=\{\edgeDirected{i_{1}}{i_{2}},\edgeDirected{i_{2}}{i_{3}},\ldots,\edgeDirected{i_{\numVerticesPath-1}}{i_{\numVerticesPath}}\}$, with $\numVerticesPath$ being the number of vertices of the path. The length of the path is defined as $\numVerticesPath-1$.
    },
}

\newglossaryentry{def:graph:adjacency}{
	name=adjacency,
	description={%
    A vertex $j$ is a predecessor of vertex $i$ iff $\edgeDirected{j}{i}\in\setEdges$.
    The set of predecessors of vertex $i$ is denoted by
    \begin{equation}
        \setPredecessors{i}=\set{j \mid \edgeDirected{j}{i}\in\setEdges}.
    \end{equation}
    A vertex $j$ is a successor of vertex $i$ iff $\edgeDirected{i}{j}\in\setEdges$.
    The set of successors of vertex $i$ is denoted by
    \begin{equation}
        \setSuccessors{i}=\set{j \mid \edgeDirected{i}{j}\in\setEdges}.
    \end{equation}
    A vertex $j$ is a neighbor to or adjacent to vertex $i$ if it is either a predecessor or a successor.
    The set of neighbors of vertex $i$ is denoted by
    \begin{equation}
        \setNeighbors{i}= \setSuccessors{i} \cup \setPredecessors{i}.
    \end{equation}
    },
    parent=def:graph,
}

\newglossaryentry{def:graph:degree}{
	name=degree,
	description={%
        The degree $\vertexDegree{i} = \lvert \setNeighbors{i} \rvert$ denotes the number of the adjacent vertices of vertex $i$. 
        The number of incoming edges called in-degree is denoted by $\vertexInDegree{i} = \lvert \setPredecessors{i} \rvert$.
        The number of outgoing edges called out-degree is denoted by $\vertexOutDegree{i} = \lvert \setSuccessors{i} \rvert$.%
    },
    parent=def:graph,
}

\newglossaryentry{def:couplingGraph}{
	name=coupling graph,
	description={A coupling graph $\graphDirected=(\setVertices,\setEdges)$ is a graph that represents the interaction between agents. Vertices represent agents and edges denote coupling objectives or constraints. A vertex from agent $i$ to agent $j$ corresponds to a coupling objective or constraint in the \ac{ocp} of agent $j$.%
    },
}

\newglossaryentry{def:matrix:Adjacency}{
	name=adjacency matrix,
	description={An adjacency matrix represents a graph with $\numAgents$ vertices in a matrix $\matAdjacency \in \set{0,1}^{\numAgents\times\numAgents}$ with entries
    \begin{equation}
        \matAdjacencyElement{ij} =
            \begin{cases}
                1 & \text{ if } \edgeDirected{i}{j} \in \setEdges \\
                0 & \text{ otherwise.}
            \end{cases}
    \end{equation}
    },
}

\newglossaryentry{def:tCompNcs}{
	name=computation time of \iac{ncs},
	description={%
        The computation time $\tCompNcs$ of \iac{ncs} is the time required for the \ac{ncs} to measure the states, formulate and solve the \ac{ocp}, apply the inputs to all agents, and communicate the required data. 
    },
}

\newglossaryentry{def:setReachable}{
	name=reachable set,
	description={%
        The reachable set of states $\setReachable$ of an agent from an initial time $t_{\text{init.}}$ to an end time $t_{\text{end}}$ is
            \begin{equation}\label{eq:setReachable}
                \setReachable_{[t_{\text{init.}},t_{\text{end}}] \mid t_{\text{init.}}} = \biggl\{ \int_{t_{\text{init.}}}^{t_{\text{end}}} \sysModelContinuous(\sysState,\sysControlInputs)dt
                \biggm| \sysState(t_{\text{init.}}) \in \setFeasibleStates(t_{\text{init.}}), \forall t: \sysControlInputs \in \setFeasibleInputs \biggr\},
            \end{equation}
        with the possible system initial states $\sysState(t_{\text{init.}})$ being bounded by its initially admissible set $\setFeasibleStates(t_{\text{init.}}) \subseteq \setRealNumbers^{\numStates}$, and the possible system control inputs $\sysControlInputs$ being bounded by its admissible set $\setFeasibleInputs \subseteq \setRealNumbers^{\numInputs}$.
    },
}

\newglossaryentry{def:conflictualDecisions}{
	name=conflictual decisions in \iac{ncs},
	description={%
        Consider two decisions made by two agents of \iac{ncs} at time step $\timestep$ with a duration $N_k$.
        They are deemed conflictual if the predicted outcome of the decisions violates the \ac{ncs}-feasibility at some point in time.%
    },
}

\newglossaryentry{def:conflictualSpace}{
	name=conflictual space of \iac{ncs},
	description={%
        In dynamic systems, the state space represents the set of all possible states the systems can occupy. 
        The conflictual space refers to a subset, or potentially the entirety, of this state space where whether decisions are conflictual is determined.
    },
}

\input{copyright.tex}

\def\BibTeX{{\rm B\kern-.05em{\sc i\kern-.025em b}\kern-.08em
  T\kern-.1667em\lower.7ex\hbox{E}\kern-.125emX}}

\begin{document}
\title{\LARGE \bf
    XP-MARL: Auxiliary Prioritization in Multi-Agent Reinforcement Learning to Address Non-Stationarity
    \thanks{This research was supported by the Bundesministerium für Digitales und Verkehr (German Federal Ministry for Digital and Transport) within the project ``Harmonizing Mobility'' (grant number 19FS2035A).}
}

\author{
    Jianye Xu$^{1}$\,\orcidlink{0009-0001-0150-2147},~\IEEEmembership{Student~Member,~IEEE},
    Omar Sobhy$^{1}$\,\orcidlink{0009-0007-4036-6110},
    Bassam Alrifaee$^{2}$\,\orcidlink{0000-0002-5982-021X},~\IEEEmembership{Senior Member, ~IEEE}
    \thanks{$^{1}$The authors are with the Chair of Embedded Software (Informatik 11), RWTH Aachen University, Germany, \href{mailto:xu@embedded.rwth-aachen.de}{\tt\footnotesize \{xu, sobhy\}@embedded.rwth-aachen.de}.}
    \thanks{$^{2}$The author is with the Department of Aerospace Engineering, University of the Bundeswehr Munich, Germany, \href{mailto:bassam.alrifaee@unibw.de}{\tt\footnotesize bassam.alrifaee@unibw.de}.}
}
    \maketitle
\thispagestyle{IEEEtitlepagestyle}
\begin{abstract}
\noindent
Non-stationarity poses a fundamental challenge in \ac{marl}, arising from agents simultaneously learning and altering their policies. This creates a non-stationary environment from the perspective of each individual agent, often leading to suboptimal or even unconverged learning outcomes. We propose an open-source framework named {\it{\frameworkName}}, which augments \ac{marl} with au\underline{x}iliary \underline{p}rioritization to address this challenge in cooperative settings.
\frameworkName is 
\begin{enumerate*}
    \item founded upon our hypothesis that prioritizing agents and letting higher-priority agents establish their actions first would stabilize the learning process and thus mitigate non-stationarity and 
    \item enabled by our proposed mechanism called action propagation, where higher-priority agents act first and communicate their actions, providing a more stationary environment for others.
\end{enumerate*}
Moreover, instead of using a predefined or heuristic priority assignment, \frameworkName learns priority-assignment policies with an auxiliary \ac{marl} problem, leading to a joint learning scheme.
Experiments in a motion-planning scenario involving \acp{cav} demonstrate that \frameworkName improves the safety of a baseline model by 84.4\% and outperforms a state-of-the-art approach, which improves the baseline by only 12.8\%.
\par\medskip
\noindent
Code: \href{https://github.com/cas-lab-munich/sigmarl}{\small github.com/cas-lab-munich/sigmarl}
\end{abstract}

\acresetall  

\section{Introduction}\label{sec:introduction}
\Ac{rl} has seen a substantial rise in both interest and application recently, primarily due to advancements in deep learning. The integration of deep learning with traditional \ac{rl} algorithms has enabled the development of agents capable of operating in complex environments, leading to significant breakthroughs in various domains, including autonomous driving \cite{shalev2016safe}, robotic control \cite{lillicrap2019continuous}, and strategy games \cite{vinyals2019grandmaster}. 

While single-agent \ac{rl} has achieved considerable success, many real-world problems inherently involve multiple agents interacting within a shared environment. Such multi-agent systems are crucial in a wide range of applications such as cooperative robots and \acp{cav} navigating traffic.
The increasing complexity and interactivity of these environments have given rise to \ac{marl}, a specialized subfield of \ac{rl} that addresses the unique challenges in multi-agent systems. However, \ac{marl} introduces additional challenges due to the dynamic interdependence among agents that single-agent \ac{rl} techniques cannot effectively handle.

One of the fundamental challenges in \ac{marl} is \textit{non-stationarity}.
In \ac{marl}, each agent considers other agents as a part of the environment. As other agents learn and update their policies continuously, the environment changes over time and becomes non-stationary from the perspective of each agent, rendering its previously learned policies less effective or obsolete. 
Therefore, the optimal policy for one agent dynamically changes, leading to a situation where each agent is essentially chasing a ``moving target'' \cite{tuyls2012multiagent}. As traditional single-agent \ac{rl} techniques often fall short in effectively handling this non-stationarity because they rely on the assumption of a stable environment, there is a pressing need to develop methods that address non-stationarity in \ac{marl}.

\subsection{Related Work}\label{sec:related}
Various works have been proposed to handle non-stationarity in \ac{marl}, see survey articles \cite{hernandez2017survey} and \cite{papoudakis2019dealing}. We detail two categories: centralized critic and opponent modeling.

\subsubsection{Centralized Critic}
Actor-critic algorithms, initially proposed for single-agent settings \cite{konda1999actor}, have been effectively adapted for multi-agent systems to address non-stationarity in \ac{marl}. The centralized critic approach is particularly appealing, leveraging centralized training with decentralized execution. During training, agents access global information that considers the states of all agents, while during execution, they operate independently using only local information. This approach stabilizes the learning process by incorporating this global information. One of the pioneering works using this approach is \acs{maddpg} \cite{lowe2017multi}, which has been widely adopted and extended by other studies, such as integrating recurrent neural networks to handle partial observability with limited communication \cite{wang2020r}, an \acs{maddpg}-based resource management scheme for vehicular networks \cite{peng2020multi}, and M3DDPG\textemdash a minimax extension for robust learning \cite{li2019robust}. Our approach can complement this category to address non-stationarity further.

\subsubsection{Opponent Modeling}
Opponent modeling addresses non-stationarity by predicting and adapting to the intentions and policies of other agents. Early approaches focused on specific domains such as Poker games \cite{billings1998opponent}, while recent research has shifted toward more generalized techniques.
Studies \cite{he2016opponent} and \cite{hong2018deep} introduced an auxiliary network that predicted opponents' actions based on observations, whose hidden layers were incorporated into a DQN to condition for a better policy. 
In contrast, \cite{raileanu2018modeling} proposed agents predicting the behavior of others using their own policies. While effective in homogeneous settings, this approach can degrade against heterogeneous opponents. In \cite{yu2022modelbased}, the authors suggested an approach that enabled agents to utilize recursive reasoning, even against similarly capable opponents. 
Other works such as \cite{zhang2010multi} and \cite{foerster2017learning} focused instead on modifying the optimization function in policy gradient methods to account for opponent learning during training. We will underscore our approach's effectiveness by comparing it with an opponent modeling approach similar to \cite{raileanu2018modeling}. 

Other promising approaches have also been proposed to address non-stationarity in \ac{marl}, including independent learning \cite{tampuu2017multiagent, foerster2017stabilising, de2020independent}, sequential learning \cite{brown1951iterative, heinrich2015fictitious, lanctot2017unified, bertsekas2021multiagent}, multi-timescale learning \cite{daskalakis2020independent, sayin2021decentralized, nekoei2023dealing}, and inter-agent communication \cite{foerster2016learning, singh2018learning, sukhbaatar2016learning}.

\subsection{Paper Contributions}\label{sec:contributions}
The main contributions of this work are threefold:
\begin{enumerate}
    \item It proposes the hypothesis that prioritizing agents and letting higher-priority ones establish actions first would mitigate non-stationarity in cooperative \ac{marl}.
    \item It introduces an open-source framework named \textit{\frameworkName}, which augments \ac{marl} with au\underline{x}iliary \underline{p}rioritization to validate the above hypothesis.
    \item It proposes a mechanism called action propagation as a key ingredient of \frameworkName, enabling lower-priority agents to condition their actions on the communicated actions of higher-priority ones, thereby providing a more predictable environment.
\end{enumerate}
Our work appears to be pioneering in addressing non-stationarity in \ac{marl} through learning-based dynamic prioritization. 
A closely related study \cite{chen2023deep}, which targeted highway merging, also incorporated priorities in \ac{marl}. However, it used handcrafted heuristics to assign priorities, limiting its application to this specific scenario and possibly resulting in inappropriate prioritization in unforeseen situations. 
In comparison, our framework extends to general \ac{marl} environments and avoids manually crafted heuristics by autonomously learning effective priority-assignment policies.

\subsection{Notation}\label{sec:notation}
A variable $x$ is marked with a superscript $x^{(\anAgent)}$ if it belongs to agent $\anAgent$. All other information is presented in its subscript, e.g., the value of $x$ at time $\timeStep$ is written as $x\ofATimeStep$. If multiple pieces of information need to be conveyed in the subscript, they are separated by commas. The cardinality of any set $\mathcal{X}$ is denoted by $|\mathcal{X}|$.
If $x$ represents a tuple, appending a new element $a$ to it is denoted by $x \gets (x, a)$.

\subsection{Paper Structure}
\Cref{sec:problem} formally formulates the problem.
\Cref{sec:framework} presents our framework as a solution. 
\Cref{sec:experiments} details experiments and discusses the limitation of our work.
\Cref{sec:conclusions} draws conclusions and outlines future research.

\section{Problem Formulation}\label{sec:problem}
A \ac{mdp} is a mathematical framework for modeling decision-making in discrete time settings with a single agent. Extending this concept, a \ac{mg} involves multiple agents, where each agent's payoff depends on the actions of all agents. In our work, we target \acp{pomg}, which allow for only partial observability of the state, requiring agents to base their decisions on incomplete information. Formally, we define a \ac{pomg} as follows (adapted from \cite{hansen2004pomg}).
\begin{definition}\label{def:pomg}
    A \ac{pomg} is defined by a tuple $(\setOfAgents, \setSystemStates, \{\actionSpace\ofAnAgent\}_{\anAgent \in \setOfAgents}, \{\observationSpace\ofAnAgent\}_{\anAgent \in \setOfAgents} ,\transitionProbFcn, \{\rewardFcn\ofAnAgent\}_{\anAgent \in \setOfAgents}, \discountFactor)$, where $\setOfAgents = \{1, \cdots, \numAgents\}$ denotes the set of $\numAgents > 1$ agents, $\setSystemStates$ denotes the state space shared by all agents, $\actionSpace\ofAnAgent$ and $\observationSpace\ofAnAgent$ denote the action and observation spaces of each agent $\anAgent \in \setOfAgents$, respectively. Let $\actionSpace \coloneqq \actionSpace \ofAgent{1} \times \cdots \times \actionSpace \ofAgent{\numAgents}$ and $\observationSpace \coloneqq \observationSpace\ofAgent{1} \times \cdots \times \observationSpace\ofAgent{\numAgents}$ denote the joint action and joint observation spaces, then $\transitionProbFcn : \setSystemStates \times \actionSpace \rightarrow \Delta(\setSystemStates) \times \Delta(\observationSpace)$ denotes the transition probability from any state $\state \in \setSystemStates$ and any given joint action $\action \coloneq (\action^{(1)},\dots,\action^{(\numAgents)}) \in \actionSpace$ to any new state $\state' \in \setSystemStates$ while receiving a joint observation $\observation \coloneq (\observation^{(1)},\dots,\observation^{(\numAgents)}) \in \observationSpace$; $\rewardFcn\ofAnAgent : \setSystemStates \times \actionSpace \times \setSystemStates \rightarrow \mathbb{\rewardFcn}$ is the reward function that determines the immediate reward received by the agent $\anAgent$ for a transition from $(\state, \action)$ to $\state'$; $\discountFactor \in [0,1)$ is the discount factor balancing the immediate and future rewards.
\end{definition}

In this setting, agents have access only to partial state of their environment. At each time step $\timeStep$, each agent \( \anAgent \in \mathcal{N} \) executes an action \( \action\ofATimeStep\ofAnAgent \) based on its partial observation \( \observation\ofATimeStep\ofAnAgent \in \observationSpace\ofAnAgent \). The environment then transitions from the current state $\state\ofATimeStep \in \stateSpace$ to a new state \( \state\ofNextTimeStep \in \stateSpace \), and the agent receives a reward \( \rewardFcn\ofAnAgent\left(\state\ofATimeStep, \action\ofATimeStep, \state\ofNextTimeStep\right) \), where \( \action\ofATimeStep \in \actionSpace \) is the joint action of all agents. The objective of agent \( \anAgent \) is to find a policy \( \policy\ofAnAgent: \observationSpace\ofAnAgent \rightarrow \Delta(\actionSpace\ofAnAgent) \), a mapping from its partial observation $\observation\ofAnAgent$ to a probability distribution $\Delta(\actionSpace\ofAnAgent)$ over its action space $\actionSpace\ofAnAgent$, such that $ \action\ofATimeStep\ofAnAgent \sim \policy\ofAnAgent (\cdot \mid \observation\ofATimeStep\ofAnAgent)$ and its expected cumulative reward\textemdash termed value function\textemdash is maximized. Note that $\sim$ denotes the sampling of an action $\action$ from a policy $\policy$.

To formally illustrate non-stationarity in \ac{marl}, we adapt the formulation in \cite{hernandez-leal2019surveya}. The value function for each agent \( \anAgent \in \setOfAgents \) is defined as:
\begin{align}
    \valueFunction_{\policy}\ofAnAgent(\state) = &\sum_{\action \in \actionSpace} \policy(\action \mid \state) \sum_{\state' \in \stateSpace} \transitionProbFcn \bigl(\state' \mid \state, \action\ofAnAgent, \action\ofOtherAgents\bigr) \times \notag \\
    &\bigl[\rewardFcn\ofAnAgent \bigl(\state, \action\ofAnAgent, \action\ofOtherAgents, \state'\bigr) \gamma \valueFunction\ofAnAgent (\state')\bigr],
\end{align}
where \( \policy(\action \mid \state) = \prod_{\anotherAgent \in \setOfAgents} \policy\ofAnotherAgent \left(\action\ofAnotherAgent \mid \state \right) \) represents the joint policy and \( \action = (\action\ofAnAgent, \action\ofOtherAgents) \) represents the joint action, with \( -i = \mathcal{N} \setminus \{\anAgent\}\) denoting the set of all agents except \( \anAgent \). The joint policy could be equivalently expressed as:
\begin{align}
     \policy(\action \mid \state) &= \policy\ofAnAgent \left(\action\ofAnAgent \mid \state\right) \cdot \underbrace{\prod_{j \in -i} \policy\ofAnotherAgent \left(\action\ofAnotherAgent \mid \state\right)} \notag\\
     &= \policy\ofAnAgent \left(\action\ofAnAgent \mid \state\right) \cdot \policy\ofOtherAgents \left(\action\ofOtherAgents \mid \state\right).
\end{align}
The optimal policy of each agent $\anAgent$, denoted as $\policyOptimal\ofAnAgent$, is the one that maximizes its value function at any state $\state$ while considering the interaction with all opponents' policies, i.e., $\policyOptimal\ofAnAgent \coloneq \arg\max_{\policy\ofAnAgent} \valueFunction\ofAnAgent_{\policy\ofAnAgent, \policy\ofOtherAgents}(\state)$.
Consequently, this optimal policy depends on the opponents' policies, which are non-stationary as the opponents learn and update their policies over time. This dynamic nature causes the optimal policy for each agent to change.

In our work, we target \textit{team \ac{pomg}}, a type of team game, where all agents share the same reward.
Formally, we formulate the standard \ac{marl} problem in a team \ac{pomg} as follows (adapted from \cite{marl-overview} and \cite{wang2002reinforcement}).
\begin{problem}[Team \ac{pomg}]\label{prob:base}
    Find a team-optimal Nash equilibrium\textemdash a joint policy $\policyOptimal \coloneq (\policyOptimal^{(1)}, \dots, \policyOptimal^{(\numAgents)})$\textemdash such that at any state $\state \in \stateSpace$,
    \begin{equation}
        \sum_{\anAgent \in \setOfAgents} \valueFunction\ofAnAgent_{\policyOptimal}(\state) \geq \sum_{\anAgent \in \setOfAgents} \valueFunction\ofAnAgent_{\policy}(\state), \forall \policy \in \policySpace
    \end{equation}
    holds, where $\policySpace$ denotes the joint policy space. 
\end{problem}

While solving \cref{prob:base} is challenging due to non-stationarity, addressing it would improve the solution quality.

\subsection{Hypothesis on Prioritization}\label{sec:hypothesis}
\begin{figure}[t]
	\centering
		\includegraphics[width=0.48\textwidth]{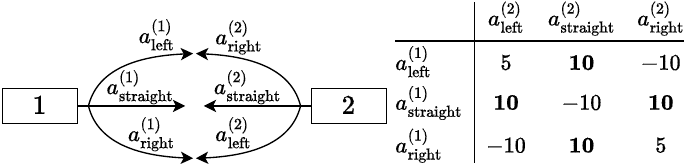}
	\caption{
        A navigation game with two agents. Each agent $\anAgent \in \{1, 2\}$ has three actions: turn left $\action_{\text{left}}\ofAnAgent$, go straight $\action_{\text{straight}}\ofAnAgent$, and turn right $\action_{\text{right}}\ofAnAgent$. Right side shows team rewards. 
    }
	\label{fig:navigation}
\end{figure}

We hypothesize that prioritizing agents and letting higher-priority agents establish their actions first can help address non-stationarity. Our hypothesis stems from the intuition that if higher-priority agents could establish their actions first and make these actions known to lower-priority agents, the former would create a more predictable environment for the latter. Consider a simple navigation game with two agents, depicted in \cref{fig:navigation}, where they learn to bypass each other. The optimal reward of 10 is issued if only one evades, while the worst reward of -10 occurs if they collide. If both evade to different sides, they receive a suboptimal reward of 5. Suppose in the previous time step, both agents tried to yield and secured a suboptimal reward. Therefore, at the current time step, they both adjust their strategies to be more aggressive and decide to go straight, which unfortunately leads to a collision and a negative reward of -10. This situation changes if one agent has a higher priority: if agent 1 acts first and makes its decision known to agent 2, agent 2 can then choose the action that maximizes the reward. 

How to assign appropriate priorities to agents remains a question. Formally, we formulate a new problem as follows.
\begin{problem}[Priority-Assignment Problem]\label{prob:stationarity}
    Find a priority-assignment policy such that at any state $\state \in \stateSpace$,   
    \begin{equation}\label{eq:priorityAssignmentProblem}
        \sum_{\anAgent \in \setOfAgents} \valueFunction\ofAnAgent_{\hat{\policy}}(\state) > \sum_{\anAgent \in \setOfAgents} \valueFunction\ofAnAgent_{\policy}(\state)
    \end{equation}
    holds, where $\hat{\policy} \in \policySpace$ and $\policy \in \policySpace$ denote the joint policy learned with and without prioritizing agents, respectively.
\end{problem}

\begin{figure*}[t]
	\centering
		\includegraphics[width=1\textwidth]{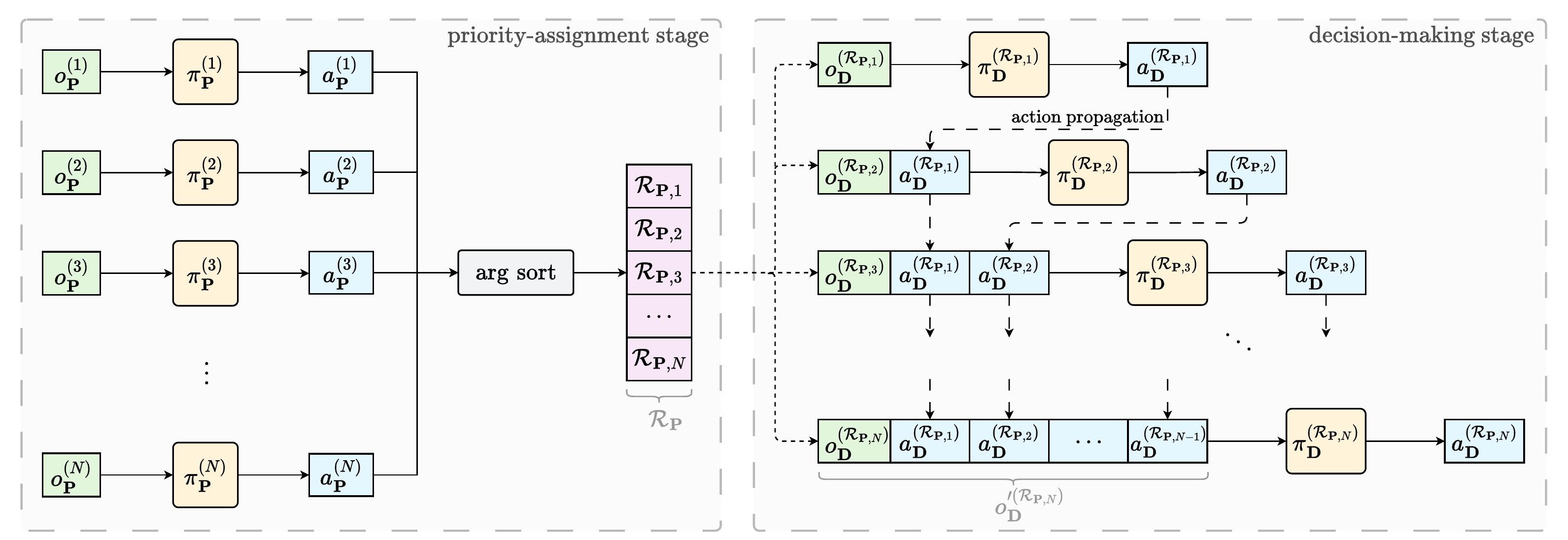}
	\caption{
        Our \frameworkName framework, time arguments omitted.
        $\observation\ofAnAgent / \action\ofAnAgent / \policy\ofAnAgent / \priRankB{\anAgent}:$ observation / action / policy / priority rank of agent $\anAgent$, $\anAgent \in \setOfAgents = \left\{1,\dots,\numAgents\right\}$.
        $\argsort$: returns the indices that sort the priority scores $(\action_{\priSubscript}\ofAnAgent)_{\anAgent \in \setOfAgents}$ in descending order.
    }
	\label{fig:overall-framework}
\end{figure*}

\section{Our \frameworkName Framework}\label{sec:framework}
This section presents our framework named \frameworkName, depicted in \cref{fig:overall-framework}.
\frameworkName consists of two stages, priority assignment and decision-making, with each corresponding to an \ac{marl} problem, thus named \textit{bi-stage} \ac{marl} problem.
We describe the bi-stage \ac{marl} problem in \cref{sec:bi-stage}, detail the two stages in \cref{sec:priority} and \ref{sec:decision}, and overview the overall framework in \cref{sec:overallFramewok}.

\subsection{Bi-Stage \ac{marl} Problem}\label{sec:bi-stage}
While heuristic priority assignment has shown effectiveness in priority-based decision-making, they are often tailored for specific scenarios or may lack the ability to handle complex situations \cite{scheffe2022increasing, chen2023deep}.
In our work, we learn priority-assignment policies by introducing an auxiliary \ac{marl} problem alongside the primary one that learns decision-making policies. This approach results in a joint learning scheme and a bi-stage \ac{marl} problem. We name the actions generated by the priority-assignment policies as \textit{priority scores} and those by the decision-making policies as \textit{decisions}. When context allows, we sometimes refer to both as \textit{actions} for simplicity.

In the remainder of the paper, variables will be annotated with subscripts to indicate their associations: \(x_\priSubscript\) for variables related to the \textbf{p}riority-assignment stage and \(x_\baseSubscript\) for those associated with the \textbf{d}ecision-making stage. 

\subsection{Priority-Assignment Stage}\label{sec:priority}
The priority-assignment stage prioritizes agents and generates a priority rank according to which agents act sequentially. This establishes a sequential decision-making scheme that serves as the backbone for the subsequent decision-making stage.

We propose \cref{alg:priority} for this stage, executed at each time step (time arguments omitted). At the beginning of each time step, for each agent $\anAgent$, the priority-assignment policy $\policy_{\priSubscript}\ofAnAgent$ generates a priority score $\action_{\priSubscript}\ofAnAgent$ based on its observation $\observation_{\priSubscript}\ofAnAgent$ (line \ref{line:rank-l3}). These scores are then appended into a tuple $\action_{\priSubscript}$, sorted in descending order, and the corresponding agents are arranged in order within another tuple \( \priRank \), which we call the \textit{priority rank} (lines \ref{line:rank-l4} to \ref{line:rank-l7}). The entire stage is visualized in \cref{fig:overall-framework} (left side).

\renewcommand{\Comment}[2][.5\linewidth]{%
  \hfill{\raggedright\scriptsize\texttt{\textcolor{black}{$\triangleright$ #2}}}}

\begin{algorithm}[t]
\caption{PriorityAssignment: Generate Priority Rank}\label{alg:priority}
\begin{algorithmic}[1]
\Input 
    joint observation: 
        $\observation_{\priSubscript} \coloneq (\observation_\priSubscript\ofAnAgent )_{\anAgent \in \setOfAgents}$, 
    joint policy: 
        $\policy_{\priSubscript} \coloneq (\policy_\priSubscript\ofAnAgent)_{\anAgent \in \setOfAgents}$
\Output priority rank of agents: $\priRank \in \setNaturalNumbers^{\numAgents}$
\State $\action_{\priSubscript} \gets ()$ \Comment{Initialize a tuple to store priority scores}
\For{$i = 1$ to $\numAgents$}
    \State $\action_\priSubscript \ofAnAgent \sim \policy_\priSubscript \ofAnAgent (\cdot \mid \observation_\priSubscript \ofAnAgent)$ \Comment{Call priority-assignment policy} \label{line:rank-l3}
    \State $\action_{\priSubscript} \gets (\action_{\priSubscript}, \action_\priSubscript \ofAnAgent)$ \Comment{Append new priority score} \label{line:rank-l4}
\EndFor
\State $\priRank \gets \argsort(\action_{\priSubscript}, \text{descending})$ \Comment{Get indices that sort a tuple} \label{line:rank-l7}
\State \Return $\priRank$
\end{algorithmic}
\end{algorithm}

\subsection{Decision-Making Stage}\label{sec:decision}
After the priority-assignment stage, the system transitions to the decision-making stage  (right side of \cref{fig:overall-framework}).

Opponent modeling approaches incorporate other agents' intentions, often by predicting their actions, which has shown effectiveness in stabilizing learning and mitigating non-stationarity \cite{billings1998opponent, he2016opponent, hong2018deep, raileanu2018modeling, yu2022modelbased, zhang2010multi, foerster2017learning}. Thus, we hypothesize that using actual actions instead of predictions could further stabilize learning. Consequently, we propose a mechanism called \textit{action propagation}, where higher-priority agents communicate their actions to lower-priority agents, enabling the latter to condition their decisions on these actions. This results in a sequential decision-making scheme where agents act one after another according to the priority rank determined during the priority-assignment stage. This structured decision-making stage reduces the complexity of interactions and enhances learning, thereby mitigating non-stationarity and improving coordination among agents.

The action propagation mechanism is enabled by the assumption that agents can communicate their actions with a communication delay of less than one time step. Note this assumption does not violate partial observability, which refers to agents' inability to fully observe the state of the environment, which is distinct from the ability to communicate specific information such as actions.

\subsection{Overview of \frameworkName}\label{sec:overallFramewok}
Integrating the above two stages leads to our framework,
\frameworkName, which enables the joint learning of the priority-assignment and decision-making policies.

\Cref{alg:overall} details the overall framework, executed at each time step (time arguments omitted). The time step begins by assigning priorities to agents and determining a priority rank $\priRank$ (line \ref{line:main-l1}). Thereafter, agents act sequentially in this ranked order. For each agent $\anAgent$, its observable higher-priority agents, denoted by $\setOfObservableHigherPriorityAgents{\anAgent}$, are identified (lines \ref{line:main-l4} to \ref{line:main-l6}). Note that we model partial observability by limiting the number of agents that each agent can observe, referred to as \textit{observable agents} henceforth. The actions of these observable higher-priority agents, denoted as $\setOfActions_{\baseSubscript, \text{prop.}}\ofAnAgent$, are propagated and appended to the agent's observation $\observation_{\baseSubscript}\ofAnAgent$, yielding a modified observation $\observation{'}_{\baseSubscript}\ofAnAgent$ (line \ref{line:main-l8}). Conditioned on this modified observation, the agent generates an action $\action_{\baseSubscript}\ofAnAgent$ using its decision-making policy $\policy_{\baseSubscript}\ofAnAgent$ (line \ref{line:main-l9}). This action is added to the joint action $\action_{\baseSubscript}$ (line \ref{line:main-l10}) to be later used by subsequent agents for action propagation. The overall framework is depicted in \cref{fig:overall-framework}.

\begin{algorithm}[t]
\caption{Overall Algorithm of \frameworkName} \label{alg:overall}
\begin{algorithmic}[1]
\setlength{\baselineskip}{1.3\baselineskip} 
\Input 
    joint observations: 
        $\observation_{\baseSubscript} \coloneq (\observation_\baseSubscript\ofAnAgent )_{\anAgent \in \setOfAgents}$ and 
        $\observation_{\priSubscript} \coloneq (\observation_\priSubscript\ofAnAgent)_{\anAgent \in \setOfAgents}$,
    joint policies: 
        $\policy_{\baseSubscript} \coloneq (\policy_\baseSubscript\ofAnAgent)_{\anAgent \in \setOfAgents}$ and 
        $\policy_{\priSubscript} \coloneq (\policy_\priSubscript\ofAnAgent)_{\anAgent \in \setOfAgents}$,
    observable agents of each agent: 
        $\{\mathcal{N}_{\text{obs.}} \ofAnAgent \}_{\anAgent \in \setOfAgents}$
\Output 
    joint action for decision-making: $\action_\baseSubscript \in \actionSpace_\baseSubscript$
\State $\priRank \gets \text{PriorityAssignment}(\observation_{\priSubscript}, \policy_{\priSubscript})$ \Comment{Call \Cref{alg:priority}} \label{line:main-l1}
\State $\action_\baseSubscript \gets ()$ \Comment{Initialize joint action} \label{line:main-l2}
\For{each $i \in \priRank$} \label{line:main-l3} \Comment{Iterate over in ranked order} 
    \State $k \gets \where(\priRank(k) \equiv i\})$ \Comment{Find index} \label{line:main-l4}
    \State $\setOfHigherPriorityAgents{\anAgent} \gets \priRank(1:k-1)$ \Comment{Higher-priority agents} \label{line:main-l5}
    \State $\setOfObservableHigherPriorityAgents{\anAgent} \gets \setOfHigherPriorityAgents{\anAgent} \cap \mathcal{N}_\text{obs.} \ofAnAgent$ \Comment{\underline{Observable} higher-priority agents} \label{line:main-l6}
    \State $\setOfActions_{\baseSubscript, \text{prop.}}\ofAnAgent \gets \{\action_{\baseSubscript}\ofAnotherAgent \mid \anotherAgent \in \setOfObservableHigherPriorityAgents{\anAgent}\}$ \label{line:main-l7}
    \State $\observation{'}_\baseSubscript \ofAnAgent \gets \observation_\baseSubscript \ofAnAgent \cup \setOfActions_{\baseSubscript, \text{prop.}}\ofAnAgent $  \Comment{Action propagation} \label{line:main-l8}
    \State $\action_\baseSubscript \ofAnAgent \sim \policy_\baseSubscript \ofAnAgent (\cdot \mid \observation{'}_\baseSubscript \ofAnAgent)$ \Comment{Call decision-making policy} \label{line:main-l9}
    \State $\action_\baseSubscript \gets (\action_\baseSubscript, \action_\baseSubscript\ofAnAgent)$ \Comment{Incrementally store actions} \label{line:main-l10}
\EndFor
\State \Return $\action_\baseSubscript$ 
\end{algorithmic}
\end{algorithm}

Note that at the end of \cref{alg:priority} and \cref{alg:overall}, the joint action of each \ac{marl} problem is fed into its environment to update the environment state and generate reward signals. However, we omit this step for simplicity.

\begin{remark}
    We present \cref{alg:priority} and \cref{alg:overall} as centralized algorithms for simplicity. They can be easily adapted to decentralized algorithms. As such, agents need to communicate their priority scores during the priority-assignment stage.
\end{remark}

\section{Experiments}\label{sec:experiments}
\begin{figure}[t]
	\centering
		\includegraphics[width=0.3\textwidth]{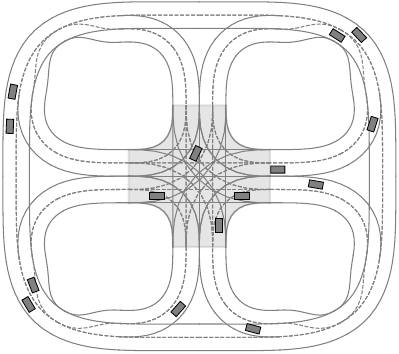}
	\caption{
        CPM scenario. Train only on the intersection (gray area) with 4 agents. Test on the entire map with 15 agents.
    }
	\label{fig:scenario}
\end{figure}

We evaluate our framework through numerical experiments within the open-source SigmaRL \cite{xu2024sigmarl}\textemdash a sample-efficient and generalizable \ac{marl} framework for motion planning of \acp{cav}. SigmaRL provides various benchmarking traffic scenarios, where agents need to cooperate to maximize traffic safety and efficiency. We select a scenario that mirrors the real-world conditions in the Cyber-Physical Mobility (CPM) Lab \cite{kloock2021cyber}, referred to as \textit{CPM scenario} thereafter. As depicted in \cref{fig:scenario}, the CPM scenario features an eight-lane intersection, a loop-shaped highway, and multiple on- and off-ramps, presenting diverse challenging traffic scenes. Our open-source repository\footnote{\href{https://github.com/cas-lab-munich/sigmarl}{\small github.com/cas-lab-munich/sigmarl}} contains code for reproducing experimental results and a video demonstration.

We detail training and testing processes in \cref{sec:training} and \ref{sec:testing}, present and interpret experimental results in \cref{sec:results} and \ref{sec:interpretation}.

\subsection{Training Process}\label{sec:training}
We conduct training in the intersection area of the CPM scenario with four vehicles, visualized by the gray area in \cref{fig:scenario}. Each vehicle is equipped with two learning agents: one for the priority-assignment policy and one for the decision-making policy. We employ \ac{mappo} \cite{lowe2017multi}\textemdash a standard \ac{marl} algorithm\textemdash to learn both policies. Each \ac{mappo} instance consists of a centralized critic and multiple decentralized actors, where each actor learns a policy for one vehicle. The input of the priority-assignment actor includes the underlying vehicle's states such as speed, observable vehicles' states such as speeds and positions, and the lane information related to centerlines and boundaries. The decision-making actor uses the same input but also includes the propagated actions of observable higher-priority vehicles. Each critic's input is a concatenation of all actors' inputs in its \ac{mappo} instance. Since vehicles are homogeneous, we enable policy-parameter sharing to enhance learning efficiency. Note that our framework can also be applied to heterogeneous multi-agent systems. We model partial observability by allowing each vehicle to observe only its two nearest neighbors.

We train five models from $\model_1$ to $\model_5$. Model $\model_1$ employs our framework, \frameworkName; Model $\model_2$ employs standard \ac{mappo} serving as a baseline; Model $\model_3$ enhances this baseline using a state-of-the-art approach, letting agents predict the actions of others using their own policies\textemdash a variant of opponent modeling inspired by \cite{raileanu2018modeling}.
Since agents share policy parameters in our experiments, they can perfectly model others, hence termed \textit{perfect opponent modeling}. Model $\model_4$ uses \frameworkName with random priorities to assess the effectiveness of the priority-assignment policies learned in Model $\model_1$. 
Model $\model_5$ also uses \frameworkName but injects noise into communicated actions, modeled by a normal distribution with a variance of \SI{10}{\percent} of the maximum action values, to test \frameworkName's robustness.

\subsection{Testing Process}\label{sec:testing}
After training, we test the models on the entire map with 15 agents rather than in the original training environment. This way, we also challenge the learned policies' ability to generalize to unseen environments. We conduct 32 numerical experiments for each model, with each lasting one minute, corresponding to 1200 time steps since each time step spans \SI{50}{\milli\second}. For each experiment, we evaluate two performance metrics: 
\begin{enumerate*}
    \item collision rate, which is the proportion of the time steps involving a collision, and
    \item relative average speed, which is the average speed of all agents relative to the maximum allowed speed.
\end{enumerate*}
These two metrics correspond to traffic safety and efficiency, respectively.

\subsection{Experimental Results}\label{sec:results}
\begin{figure}
    \centering
    \begin{subfigure}[t]{1\linewidth}
        \centering
        \includegraphics[scale=0.9]{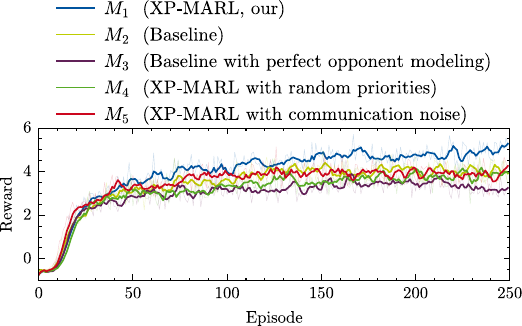}
        \caption{Episode mean reward during training. Smoothed with a sliding window spanning five episodes.}\label{fig:reward}
    \end{subfigure}

    \medskip
    
    \begin{subfigure}[t]{0.96\linewidth}
        \centering
        \includegraphics{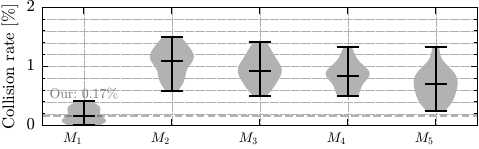}
        \caption{Collision rate during testing.} \label{fig:collisionRate}
    \end{subfigure}

    \medskip

    \begin{subfigure}[t]{0.96\linewidth}
        \centering
        \includegraphics{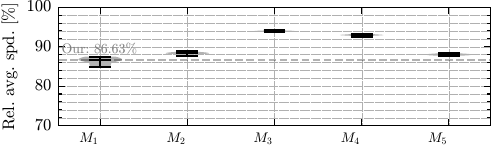}
        \caption{Relative average speed during testing.}\label{fig:averageSpeed}
    \end{subfigure}
    \caption{
        Training curves and testing results of 32 experiments.
    }\label{fig:experiments}
\end{figure}

\Cref{fig:reward} depicts the episode mean reward during training.
Model $\model_1$ secures the highest reward, indicating the best learning efficiency. This suggests that our framework effectively mitigates non-stationarity in our experiments. All other four models demonstrate similar learning efficiency, with Model $\model_3$ exhibiting slightly worse performance.

\Cref{fig:collisionRate} presents the collision rate during testing. The baseline model, $\model_2$, exhibits a median collision rate of \SI{1.09}{\percent}.
Remarkably, Model $\model_1$, trained with our \frameworkName, lowers this rate by \SI{84.4}{\percent}, outperforming Model $\model_3$, which uses perfect opponent modeling and achieves an improvement of only \SI{12.8}{\percent}. The unsatisfactory performance of Model $\model_3$ may result from its inability to handle non-stationarity as effectively as our framework, which we will elaborate on in \cref{sec:interpretation}. Further, Model $\model_4$ improves the baseline by \SI{22.0}{\percent}, suggesting that incorporating our prioritization scheme in \ac{marl} can mitigate non-stationarity, even with random priorities. Additionally, Model $\model_1$ surpasses Model $\model_4$, demonstrating effective learning of priority-assignment policies. Moreover, even with \SI{10}{\percent} communication noise, Model $\model_5$ still reduces the collision rate of the baseline by \SI{35.8}{\percent} and outperforms Model $\model_3$, validating the robustness of our approach.

\Cref{fig:averageSpeed} depicts the average speeds, reflecting traffic efficiency. With reduced non-stationarity, the agents in Model $\model_1$ have learned to brake properly to avoid collisions rather than recklessly cruise at a high speed. As a result, Model $\model_1$ achieves the lowest traffic efficiency. Nevertheless, it only marginally reduces traffic efficiency by \SI{2.2}{\percent} compared to the baseline, while improving its safety by \SI{84.4}{\percent}.


\begin{figure}[t]
    \centering
    \subfloat[\centering Approaching an on-ramp.]{{\includegraphics[width=0.12\textwidth]{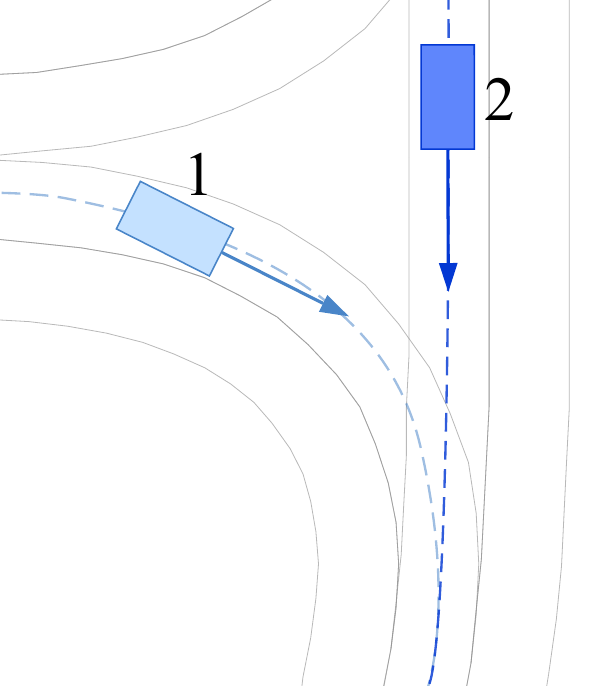}}}%
    \hfill
    \subfloat[\centering Switching priorities.]{{\includegraphics[width=0.12\textwidth]{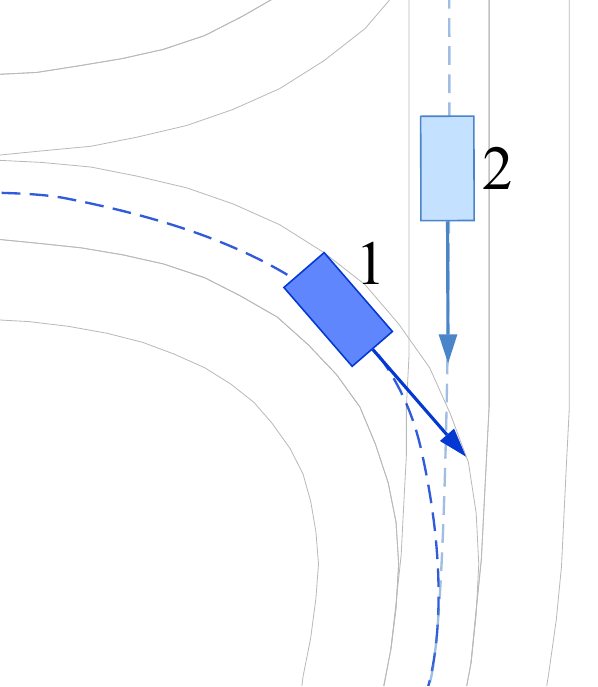}}}%
    \hfill
    \subfloat[\centering Agent 2 yields.]{{\includegraphics[width=0.12\textwidth]{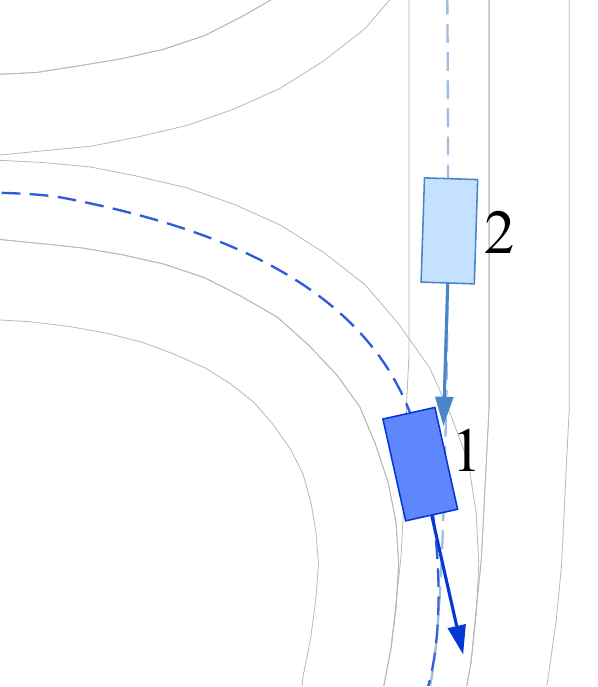}}}%
    \hfill
    \subfloat{\includegraphics[width=0.06\textwidth]{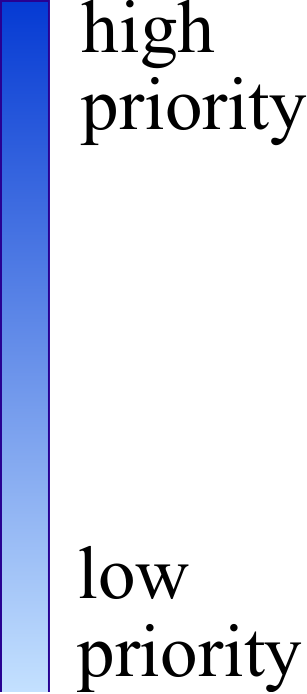}}%
    \caption{Two agents avoiding a collision at an on-ramp by dynamically switching their priorities.}
    \label{fig:ramp}
\end{figure}

\subsection{Interpretation of the Experimental Results}\label{sec:interpretation}
Recall the navigation example in \cref{fig:navigation}. Assume agents can perfectly predict the actions of their teammates, as they do in Model $\model_3$. They may still fail to secure the optimal reward: if at the current time step, they follow the same aggressive policy. Both agents will predict that the other will go straight and decide to evade, leading to either a collision if they evade to the same side or a suboptimal reward if they evade to different sides. In contrast, using our \frameworkName, even with random priorities, allows them to achieve the optimal reward in all cases. This may explain why Models $\model_1$ and $\model_4$ outperform Model $\model_3$ in terms of safety in our experiments. 

\Cref{fig:ramp} exemplifies two agents approaching an on-ramp during our experiments. Before entering the on-ramp, agent 1 holds a lower priority than agent 2. Upon entering the on-ramp, they switch priorities, making agent 2 responsible for collision avoidance. This way, agent 2 sacrifices its individual short-term reward for a better team benefit.

\section{Conclusions}\label{sec:conclusions}
In our work, we proposed an open-source framework named \textit{\frameworkName}, which augmented \ac{marl} with au\underline{x}iliary \underline{p}rioritization and mitigated the non-stationarity challenge in \ac{marl}. It was founded upon our hypothesis that prioritizing agents and letting higher-priority agents establish their actions first would mitigate this challenge. It incorporates a mechanism called action propagation, which propagates the actions of high-priority agents, thus stabilizing the environment for lower-priority agents. Instead of using predefined or heuristic priority assignment, it jointly learns priority-assignment policies and decision-making policies. We evaluated its effectiveness with SigmaRL\textemdash an \ac{marl}-based motion-planning framework for \acp{cav}. It improved the safety of a baseline model by \SI{84.4}{\percent} and outperformed a state-of-the-art approach employing perfect opponent modeling, which improved the baseline by only \SI{12.8}{\percent}. Moreover, even with \SI{10}{\percent} communication noise, it still maintained an improvement of \SI{35.8}{\percent}, demonstrating its robustness. These results validated our hypothesis and suggested that prioritizing agents might be a promising approach to mitigating the non-stationarity challenge in \ac{marl}.

The sequential planning scheme of our \frameworkName may result in prolonged idle times for lower-priority agents, since they need to wait for the communicated actions of higher-priority agents, potentially limiting its applicability to large-scale systems. 
In future work, we plan to integrate our previous works \cite{scheffe2024limiting} and \cite{scheffe2023reducing}, which leverage graph theory and reachability analysis, into \frameworkName to address this limitation.
\bibliographystyle{IEEEtran}
\bibliography{references.bib}

\end{document}